\documentclass[11pt]{article}

\usepackage{graphicx}
\usepackage{setspace}
\usepackage[cmex10]{amsmath}
\usepackage[tworuled,linesnumbered,noend,noline]{algorithm2e}
\usepackage[absolute,showboxes]{textpos}
\usepackage[left = .56in, top=.56in,right=.56in,bottom=0.56in,nohead,nofoot]{geometry}

\begin{document}
\title{GraATP: A Graph Theoretic Approach for Automated Theorem Proving in Plane Geometry}



\author{Mohammad Murtaza Mahmud, Swakkhar Shatabda and Mohammad Nurul Huda \\Department of Computer Science and Engineering, United International University\\House \# 80, Road \# 8A, Dhanmondi, Dhaka-1209, Bangladesh\\Email: mohammadmurtazamahmud@gmail.com, swakkhar@cse.uiu.ac.bd, mnh@cse.uiu.ac.bd}


%


\maketitle
\doublespacing
\begin{abstract}
Automated Theorem Proving (ATP) is an established branch of Artificial Intelligence. The purpose of ATP is to design a system which can automatically figure out an algorithm either to prove or disprove a mathematical claim, on the basis of a set of given premises, using a set of fundamental postulates and following the method of logical inference. In this paper, we propose GraATP, a generalized framework for automated theorem proving in plane geometry. Our proposed method translates the geometric entities into nodes of a graph and the relations between them as edges of that graph. The automated system searches for different ways to reach the conclusion for a claim via graph traversal by which the validity of the geometric theorem is examined. 

\end{abstract}

\section{Introduction}
In a geometric theorem, basically we are given a set of hypotheses which we have either to prove or disprove. Depending on these hypotheses, we figure out the whole geometric system. A list of fundamental postulates and previously proven theorems, are known. They are used to infer the related geometric facts from the given hypotheses. These derived geometric facts which have been discovered so far are used further to derive more geometric facts until the conclusion is reached about the claim of the given theorem. Alternatively, it is possible to figure out the geometric facts which must be true if the claim is to be true. To do so, one needs to use the fundamental geometric postulates and apply the process of logical inference. Consequently, the theorem-prover infers what other geometric facts are required to be true if the previously derived geometric facts are to be remained satisfied. The process is carried on until the theorem-prover discovers that the required facts for the validity of the final claim are given as the hypotheses of the theorem. An `intelligent thinker' thinks in both ways to generate a particular algorithm to prove a theorem. 
\textit{Automated Theorem Proving (ATP)} is enabling a machine (computer) to figure out an algorithm to prove a given theorem by the mechanization of the above mentioned process.

ATP has been established as a branch of Artificial Intelligence for several decades. In 1954 Martin Davis, an American Mathematician programmed Presburger’s algorithm \cite{davis1957computer}. Later Allen Newell, Herbert A. Simon and J. C. Shaw developed Logic Theory Machine around 1955-56 \cite{newell1956logic}. In 1959 they created General Problem Solver (G.P.S.) \cite{newell1963guide} which was able to solve any symbolic problem. Gelernter, J. R. Hanson and D.W. Loveland worked on geometric theorem proving implementing traditional proof method \cite{gelernter1960empirical}. However, their method suffers difficulties of the explosion of the search space. Later Wen-Tsun Wu developed an algebraic method \cite{elias2006automated} which could prove geometric theorems more efficiently, but this method involves lots of calculations with polynomials which make the proof hardly readable. Chou, Gao and Zhang \cite{chou1994machine} developed `area method' which is able to produce short and readable proofs of geometric theorems. In his paper, David A. Plaisted \cite{plaisted2014automated} reviewed different techniques of ATP. Among these techniques are: propositional proof procedures \cite{malik2009boolean,moskewicz2001chaff}, first order logic \cite{fitting2014possible}, clause linking \cite{lee1992eliminating}, instance-based procedures \cite{plaisted2000ordered}, model evolution \cite{baumgartner2014model}, modulo theories \cite{de2011satisfiability}, unification and resolution \cite{lassez1991computational} and combined systems \cite{bridge2013case,armando2009new}. In another paper, Joran Elias \cite{elias2006automated} discussed Wu’s method on geometric theorem proving.

There are two broad categories of techniques to prove a geometric theorem. They are: Euclidean Logical Inference methods \cite{fu2014geometry} and Cartesian Algebraic methods \cite{franova2014cartesian}. The former method uses logical inference to reach at conclusion from a set of premises. On the other hand the later method converts a given set of premises into a set of algebraic equations and then solves those equations for unknown parameters. In this paper, we propose GraATP, an ATP combining both algebraic method (Cartesian Analytical Geometry) and logical inference method (Euclidian geometry) to prove geometric theorems. Our proposed method translates the geometric entities into nodes of a graph and the relations between them as edges of that graph. The automated system searches for different ways to reach the conclusion for a claim via graph traversal by which the validity of the geometric theorem is examined.

Rest of the paper is organized as follows: first we discuss the preliminaries required to figure out a geometric structure in Section~\ref{preli}. We describe Cartesian analytical geometry and traditional Euclidean proof using logical inference method in Section~\ref{sec21} and Section~\ref{sec22} respectively. In Section~\ref{secMet}, we propose our method combining these two methods to prove geometric theorems. Finally, we conclude the paper with an outline of the future work in Section~\ref{secCon}.

\section{Preliminaries\label{preli}}
To define a geometric system, we use four elementary concepts of geometry: point, straight line, angle and circular arc. Usually, we choose a point and a line passing through the point as an initial reference. Position of a point is specified by a distance from another previously defined point along a particular straight line. Orientation of a line is specified by the angle made by it with another previously specified line and the point of intersections between the lines. A circular arc is specified by the position of its central point and it radius. For example, following steps are required to derive a parallelogram in \figurename~\ref{f1}:
\begin{figure}
\begin{center}
\includegraphics[scale=0.5]{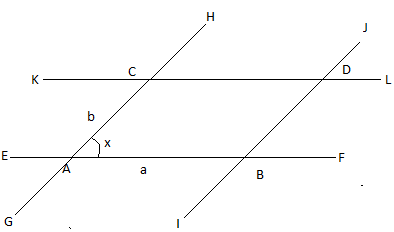}
\end{center}
\caption{A geometric system of lines and points.}
\label{f1}
\end{figure}
\begin{enumerate}
\item $A$ is a reference point
\item $EF$ passing through $A$ is a reference line
\item Line $GH$ passes through $A$, angle $\angle FAH = x$
\item $C$ is a point on $GH$ where $AC = b$
\item Line $KL$ passes through $C$, angle $\angle KCG = x$
\item Line $IJ$ passes through $B$, angle $\angle FBJ = x$
\item $D$ is the intersection of the line $KL$ and $IJ$ is determinable since $KL$ and $IJ$ are specified
\item \{$AC, CD, DB, B$A\} is the parallelogram
\end{enumerate}

Once we able to figure out a complete geometric structure, we can explore different dimensions (lengths of the lines, angles between lines, etc) of the structure. Hence, we can test whether a certain claim is true or false knowing these dimensions.

\subsection{Cartesian Method\label{sec21}}
In Cartesian method, geometry is combined with algebra. Two axes, perpendicular to each other and their point of intersection, i.e. origin, are specified. A point on a plane is specified by pair of coordinates which are the distances of the point from the origin along the axes. Curves and straight lines are specified by algebraic equations. Solving these equations unknown dimensions are worked out. Finally, facts to be proven are verified.

\begin{figure}[h]
\begin{center}
\includegraphics[scale=0.5]{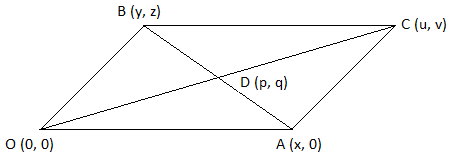}
\end{center}
\caption{A geometric system of a parallelogram.}
\label{f2}
\end{figure}

Let’s consider the following example from \cite{elias2006automated}. We have to prove that diagonals of a parallelogram bisect each other. Please see \figurename~\ref{f2}. Here, the hypotheses are - i) $OACB$ is a parallelogram $\implies OB||AC, OA||BC$, $OC$ and $AB$ are diagonals, ii) $D$ is the point of intersection of $AB$ and $OC$. First, we have to decompose these statements into a couple of equations.

As mentioned earlier, we have to specify the points of our interest- $O, A, C, B$ and $D$ each with two coordinates. Let $O, A$ and $B$ are denoted by $(0, 0), (x, 0)$ and $(y, z)$ respectively. Here, $x, y$ and $z$ are arbitrary parameters what we have chosen. Once we choose $x, y$ and $z,$ the coordinates of $C$ and $D$ become fixed depending on $(x, y, z)$ according to the hypotheses. Let us assume that coordinates of $C$ and $D$ be $(u, v)$ and $(p, q)$.
Since $OB$ and $AC$ are parallel to each other, their slopes are equal too. Hence we get,
\begin{equation}
uz-zx=v	y
\label{eq1}
\end{equation}

On the other hand, $OA$ and $BC$ are parallel to each other, their slopes are equal too. Hence we get,
\begin{equation}
v=z
\label{eq2}
\end{equation}

We can work out $u$ and $v$ in terms of $x, y$ and $z$   by solving Equation~\ref{eq1} and Equation~\ref{eq2}. Finally, we find out the length of $OD, DC, BD$ and $AD$ by using Pythagoras theorem. If we can show, $OD=DC$ and $BD=AD$ then the theorem is proved. 

\subsection{Euclidean Logical Inference Method \label{sec22}}
In logical inference method, a set of axioms, previously proved theorems and hypotheses are used to discover the relationship among different entities (lengths of line segments or arcs, positions of points, amount of angles and equalities or similarities of finite regions like triangles) of a geometric structure. These relationships are used to proceed further to infer relationship among different other entities from the previously derived relationships. This process continues until the relationship between two particular entities of interest is discovered. Let's think about the previous example: diagonals of a parallelogram bisect each other.

We have to discover the relationship between the entities (here length of two line segments): $OD$ and $CD$ as well as $BD$ and $AD$. First of all, we will find out relations exploiting the hypotheses. Since $OACB$ is a parallelogram, $(OB, AC)$ and $(OA, BC)$ are opposite sides, they are parallel and equal to each other. $BA$ is the common sector of $OB$ and $AC$. Hence the $\angle OBA$ is equal to the $\angle BAC$. Here, we used a previously discovered theorem: if a line intersects two parallel lines then the alternate angles created in the points of intersection are equal. Similarly, we find out the relationship between $\angle BOC$ and $\angle OCA$. Since $D$ is a point on $AB$, angle $\angle OBA =  \angle OBD$. Similarly, $\angle BAC = \angle DAC$. Again $D$ is a point on $OC$. Hence, $\angle BOC =  \angle BOD$ and  $\angle OCA = \angle DCA$. Now in  $\triangle BOD$ and $\triangle ACD, OB = AC,  \angle OBD =  \angle DAC$ and  $\angle BOD =  \angle ACD$. Therefore,  $\triangle BOD$ and  $\triangle ACD$ are equal. Here, we used another previously discovered theorem: if two triangles have a side of equal length and two adjacent angles of equal amount each, then the triangles are equal. $OD$ is the opposite side of the $\angle OBD$ and $CD$ is the opposite side of the  $\angle CAD$. Since  $\triangle BOD$ and  $\triangle ACD$ are equal and $\angle OBD= \angle CAD \implies OD = CD$. Similarly, $BD = AD$. This is the desired relationship to prove the theorem. Our process of searching information on how different entities are related with each other throughout the geometric structure stops here.

\section{GraATP: Our proposed ATP Framework\label{secMet}}
In the previous section, we discussed two manual approaches for geometric theorem proving. If we compare between two ways, at a first glance, Cartesian algebraic method seems complicated than the logical inference method. Algebraic method is mechanical, all we have to do is to fix the position coordinates of some particular points, discover equations of straight lines or curves appearing in the geometric structure and find out the coordinates of other points as functions of the co-ordinates of the previously fixed points. When we know all dimensions of the structure we test whether the final claim is true or false. On the other hand, Euclidean logical inference method requires more heuristic knowledge, i.e. more `intelligence' to discover the hidden relationship among different entities of the structure. Prover's skill to observe the geometric structure, and retrieve the previously discovered theorems, related to the problem, from the memory, play important role here. Moreover, whether the searching process (the process of discovering relationship among the entities) approaches towards the goal (testing the relationship which is supposed to be proven) depends on the prover’s intuition. By comparing the two methods, we can conclude that the automation of Cartesian method is easier than the logical inference method.

\begin{figure}[h]
\begin{center}
\includegraphics[scale=0.5]{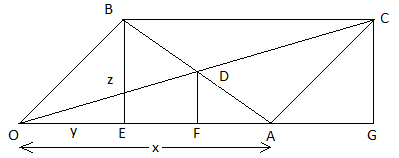}
\end{center}
\caption{A geometric system of a parallelogram.\label{f3}}
\end{figure}

Here, we propose a primitive approach of finding out an algorithm to prove a geometric theorem in an automated way. There are several previously proposed ways: Wu’s method \cite{elias2006automated}, Area method \cite{chou1994machine}, etc. Our goal is to build up a framework of finding an algorithm that resembles the way in which we the human or intelligent theorem prover thinks to prove a theorem. Let’s discuss the previous example again in a different way. Consider the geometric system in \figurename~\ref{f3}.

Our hypotheses are as follows:
\begin{enumerate}
\item $OA = x$
\item $E$ lies on $OA$
\item $BE$ is perpendicular to $OA$
\item $OE$ = $y$
\item $EB$ = $z$
 \item $OB||AC$
\item $OA||BC$
\item $D$ lies on $AB$
\item $D$ lies on $OC$
\item $DF$ is perpendicular to $OF$
\item $A$ lies on $OG$
\item $CG$ is perpendicular to $OG$

\end{enumerate}

We have to show that $OD=CD$ and $BD=DA$. Here, we get a unique geometric structure for a unique set of the parameters $(x, y, z)$. Our next goal is to explore the geometric structure to express all of the dimensions (length of the segments of lines) as functions of these three parameters $x, y$ and $z$. When $OD, CD, BD$ and $DA$ can be expressed as functions of $x, y$ and $z$, then the process of exploration stops. If $OD=CD$ and $BD=DA$, then the claim is proved. 

A possible sequence to work out different dimensions are as follows:
\begin{enumerate}
\item Find $CG$. $CG = BE=z$ (exploiting the fact that $BC||OA$ and $G$ lies on the extension of $OA$) 
\item Find $\frac{CG}{AG} (= \frac{BE}{OE} =\frac{z}{y})$ (exploiting the fact that $\triangle OBE$ is similar to the $\triangle ACG$)
\item Find $AG$, since we know $CG$ and the ratio  $\frac{CG}{AG}$. 
\item Find $OG$. $OG = OA + AG$
\item Find $\frac{DF}{OF} (=\frac{CG}{OG})$ exploiting the fact that $\triangle DFO$ and $\triangle CGO$ are similar.
\item Find $AE$. $AE = OA-OE$
\item Find $\frac{AF}{DF}$ which equals to $\frac{AE}{BE}$ ( $\triangle ADF$ and $\triangle ABE$ are similar)
\item Express $AF = OA-OF$
\item Find $DF$ and $OF$ using the ratio $\frac{DF}{OF}$and $\frac{OA-OF}{DF}$
\item Find $OD, OD =\sqrt{OF^2+DF^2}$
\item Find $CD: CD =\sqrt{(OG-OF)^2+(CG-DF)^2}$
\item Check whether $OD = CD$
\end{enumerate}

Here, if $OD$ and $CD$ are equal then the theorem is proved. In the same way we can check whether $AD$ and $BD$ are equal or not.

\begin{figure}[h]
\begin{center}
\includegraphics[scale=0.5]{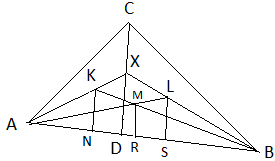}
\end{center}
\caption{A geometric system of a triagle. \label{f4}}
\end{figure}

Now, we present another example, more complicated than the previous one. Please see \figurename~\ref{f4}. Let $\triangle ABC$ is a triangle with  $\angle BCA = 90^o$ and let $D$ be the foot of the altitude from $C$. Let $X$ be a point in the interior of the segment $CD$. Let $K$ be the point on the segment $AX$, such that $BK=BC$.  Similarly, let $L$ be the point on the segment $BX$ such that $AL = AC$. Let $M$ be the point of intersection of $AL$ and $BK$. We have to show that, $MK = ML$\footnote{This problem is taken from the International Mathematics Olympiad 2012 http://www.imo-official.org/problems/IMO2012SL.pdf}.

Let's rephrase the hypotheses in the following way:
\begin{enumerate}
\item $AD = a$
\item $CD\perp AB$ and $CD = h$
\item $A,C$ are added by a line segment
\item $CB \perp AC$ at $C$
\item $B$ lies on the extension of the line $AD$
\item $X$ lies on $CD$ where $XD = q$
\item $A, X$ are added by a line segment
\item $B, X$ are added by a line segment
\item $K$ lies on $AX$ such that $BK = BC$
\item $L$ lies on $BX$ such that $AL = AC$
\item $M$ is the point of intersection between $BK$ and $AL$

\end{enumerate}

Here, we get a unique geometric structure for a unique set of the parameters $(a, h, q)$. Next goal is to explore the geometric structure to express all of the dimensions (length of the segments of lines) as functions of these three parameters $a, h$ and $q$. When we will be able to express $LM$ and $MK$ in terms of $(a, h, q)$ then the process of exploration stops. If the two functions are equal then the claim if proved. 

Our proposed method GraATP will find out a sequence of the dimensions (which need to be worked out in terms of $(a, h, q)$ of this geometric structure starting from $(AD=a, CD=h, XD=q)$ to $(LM, KM)$. To locate the points $K, M$ and $L$ we draw $KN, MR$ and $LS$ perpendicular to $AB$. A possible sequence of working out the dimensions is:
\begin{enumerate}
\item Find $AC: AC =\sqrt{a^2+h^2}$  
\item Find $BD$ (exploiting the similarity between  $\triangle ABC$ and  $\triangle ADC$)
\item Find $BC$ (exploiting the similarity between  $\triangle ABC$ and  $\triangle ADC$)
\item Find $AX: AX =\sqrt{a^2+q^2}$  
\item Find $BX: BX =\sqrt{BD^2+q^2}$  
\item Find $KN$ and $AN$ (exploiting the similarity between  $\triangle AKN$ and $\triangle AXD$, and applying Pythagoras theorem in $\triangle BKN$)
\item Find $LS$ and $AS$ (exploiting the similarity between $\triangle BXD$ and $\triangle BLS$, and applying Pythagoras theorem in $\triangle ALS$)
\item Find $BN = AB-AN$
\item Find $AS = AB-BS$ 
\item Find $MR$ and $AR$ (exploiting the similarity between triangles $(\triangle BMR, \triangle BKN)$ and ($\triangle AMR, \triangle ALS$)
\item Find $KM$ ($KM^2 = (AR-AN)^2 + (KN-MR)^2$)
\item Find $ML(ML^2=(AS-AR)^2+(LS-MR)^2$
\item Check whether $KM = ML$
\end{enumerate}

By observing the commonalities between the two above mentioned techniques we can formulate a general way to find a theorem proving algorithm as follows:
\begin{enumerate}
\item Specify a set of parameters by means of which the geometric structure can uniquely be constructed
\item Find out different dimensions of the structure by means of the predefined parameters [to do so we use basically similarity between triangles and Pythagoras theorem]
\item Continue step 2 until the dimensions of a set of particular elements are found
\item Check whether the claim is true
\end{enumerate}

The whole process can be represented as the formation of a graph and traversing through the graph. We can represent different dimensions (length of line segment, angle and circular arc-length) and the functions of dimensions (for example, ratio of two line segments) as nodes of the graph. Using the hypotheses of the theorem, we discover the relationships among the dimensions. If we can work out the node $A$ from node $B$ then we draw a directed edge from $B$ to $A$. In the evolutionary process of the formation of the graph, we put the nodes showing the dimensions which we choose as parameters. In the parallelogram example, these dimensions are $OA(=x), OE(=y)$ and $BE(=z)$.
\begin{figure}[t]
\begin{tabular}{cc}
\includegraphics[width=.2\textwidth]{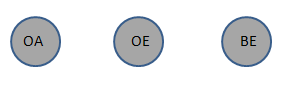}&\includegraphics[width=.2\textwidth]{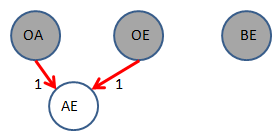}\\
(a)&(b)\\
\includegraphics[width=.2\textwidth]{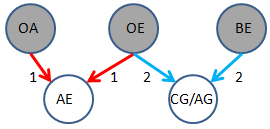}&\includegraphics[width=.2\textwidth]{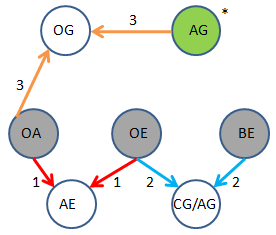}\\
(c)&(d)\\
\includegraphics[width=.2\textwidth]{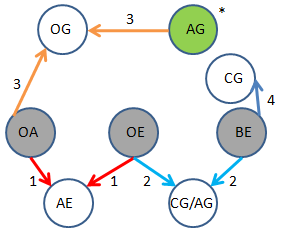}&\includegraphics[width=.2\textwidth]{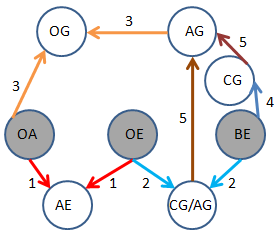}\\
(e)&(f)\\
\end{tabular}
\caption{Steps of the evolutionary process for the formation of the graph. \label{f5}}
\end{figure}

\figurename~\ref{f5} (a) shows the initial step. The color gray denotes the nodes that are the chosen as parameters; no other dimensions are required to know to find out their values. Hence, edges from other nodes will not be incident on them. Now, using the hypotheses we will see which dimensions are closely connected to these three dimensions and include them in the graph. Since $E$ lies on $OA, AE = OA-OE$. We can find out $AE$ from $OA$ and $OE$. In the second step, we include another node $AE$ (shown in \figurename~\ref{f5} (b)). Also we include two edges one from $OA$ to $AE$ and another from $OE$ to $AE$; and we draw them with same color (red) and label them with number 1 to indicate that the set of dimensions $\{OA, OE\}$ is required to be known to find out $AE$. A same node can be found out by knowing different sets of dimensions. In that case, we would choose different colors and labels.

In the next step, we exploit the similarity between $\triangle OBE$ and $\triangle ACG$ to discover more relations: $\frac{CG}{AG}=\frac{BE}{AG}$. Therefore, we can include another node, this time a ratio of dimensions, $\frac{CG}{AG}$ (\figurename~\ref{f5} (c)). Blue edges labeled with number 2 come out from the nodes $OE$ and $BE$ and they are incident on the node $\frac{CG}{AG}$. Next, $A$ lies on $OG$. Hence $OG = OA+AG$. We include nodes $AG$ and $OG$. We draw two edges, one from $OA$ and another one from $AG$ to $OG$. They are labeled with number 3. The dimension $AG$ is not a parameter and still no edges are incident on it from any other node which can be represented as a function of the parameters ${OA, OE, BE}$. That's why we have made it lime colored (\figurename~\ref{f5} (d)) and put an asterix mark on it. It means that we have to discover more node(s) from which edge(s) will come out to meet $AG$ and connect $AG$ with the nodes which have already been discovered. In the next step, we use the fact that $BC||OG$ to decide that $BE = CG$. Therefore, we add another node $CG$ and draw an edge from $BE$ to $CG$ (\figurename~\ref{f5} (e)). Now, we can find out $AG$ from $CG$ and the ratio $\frac{CG}{AG}$. So we draw two edges: one from $\frac{CG}{AG}$ to $AG$ and another from $CG$ to $AG$ (\figurename~\ref{f5} (f)). The node $AG$ is connected with the discovered nodes, so its color becomes white now and the asterix mark is dropped.

The process continues until:
\begin{enumerate}
\item A connected graph is formed containing the parameter-nodes ($OA, OE,BE$) and the destination-nodes ($CD, OD$),
\item There exists no node having no incoming edges except for the parameter-nodes. As for example in step 4 the node $AG$ was included. There was no edge which is directed from other node to $AG$. Also $AG$ is not one of the parameter-nodes like $OA$, $OE$ and $BE$. Therefore the process of forming the graph continues.
\end{enumerate}
The algorithm is given in Algorithm~\ref{alg1}.

\begin{algorithm}
\DontPrintSemicolon
$H:$ set of hypotheses\;
$R:$ set of conclusions\;
$D:$ set of dimensions \;
$P=$ create a set of unique parameters\;
$E\leftarrow\phi$\;
$V\leftarrow\phi$\;
$G=\langle V,E \rangle$\;
\For{each $p\in P$}
{
	create a node $u$\;
	$V=V\cup {u}$\;
}
\For{each $r\in R$}
{
	create a node $u$\;
	$V=V\cup {u}$\;	
}
\While{$D \neq \phi$}
{
	create node $u$ for the next close dimension $d\in D$\;
	\For{each $v \in V$ that is related to $u$}
	{
		add a directed edge $(u,v)$ or $(v,u)$\;
	}	
	remove $d$ from $D$\;
}
\If{$G$ is not connected}
{
	\Return null\;
}
\Else
{
	\Return $G$\;
}

\caption{GraATP (H,R)}
\label{alg1}
\end{algorithm}

\begin{figure}[h]
\begin{tabular}{cc}
\includegraphics[width=.25\textwidth]{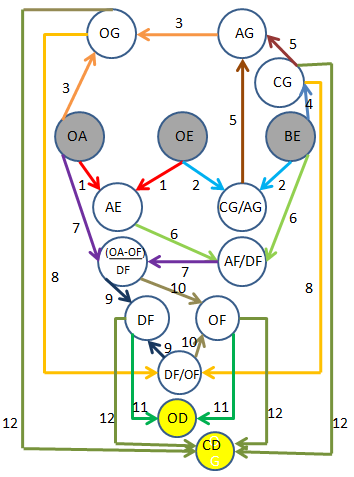}&\includegraphics[width=.2\textwidth]{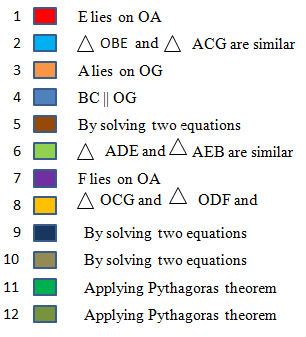}\\
\end{tabular}
\caption{The complete graph of proving the theorem on parallelogram. How the edge-relations between nodes are discovered, are also mentioned.\label{f6}}
\end{figure} 

\begin{figure}[h]
\begin{center}
\includegraphics[width=.3\textwidth]{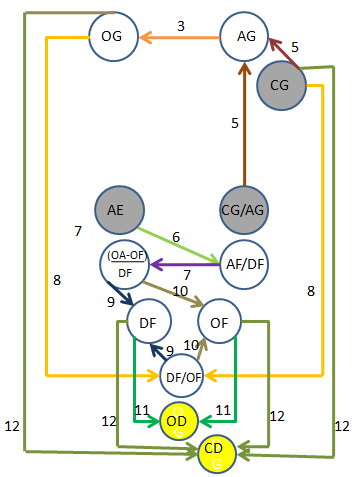}
\end{center}
\caption{First step of the topological ordering algorithm.\label{f7}}
\end{figure}

\figurename~\ref{f6} shows the complete graph to reach $OD$ and $CD$ from ${OA, OE, BE}$. Now we will apply standard topological ordering algorithm to find out the sequence of steps of the theorem proving algorithm. First, we will enlist the nodes having no incoming edges. They are the parameter nodes: $OA, OE$ and $BE$. Next, we delete these enlisted nodes and the edges adjacent of them as shown \figurename~\ref{f7}.

After that, we look for the nodes having no incoming edges in the new graph. They are $CG/AG, CG$ and $AE$. We delete them and their adjacent edges from the graph. We proceed in this way until we reach the destination vertices $CD$ and $OD$. Therefore, the topological order of the nodes is: $OA, OE, BE, CG/AG, CG, AE, AF/DF, AG, OG, (OA-OF)/DF, DF/OF, DF, OF, CD, OD$. The topological sorting algorithm is given in Algorithm~\ref{alg3}.

\begin{algorithm}
\DontPrintSemicolon
$A=\phi$\;
$L\leftarrow$ set of all nodes with indegree = 0\;
\While {$L\neq \phi$}
{
$u\leftarrow L.\textsf{extractNode()}$\;
$A.\textsf{addToLast}(u)$\;
\For{each $v\in Adj[u]$}
{
 $E= E-(u,v)$\;
}
}
\If {$E\neq \phi$}
{
	\Return null
}
\Else
{
	\Return $A$
}

\caption{Topological Ordering ($G=\langle V,E\rangle$)\label{alg3}}
\label{alg1}
\end{algorithm}

\section{Conclusion\label{secCon}}
So far we have discussed how to translate a geometric structure, which is uniquely configured by setting a set of parameters, to a graph and how to traverse through the graph to find out a sequence of steps performing which the theorem can be proven. There are several mechanical methods of proving geometric theorems which have already been proposed, e.g. Wu’s method \cite{elias2006automated}, Area method \cite{chou1994machine}, and so on. The purpose of this work is to resemble the way in which human thinks, perhaps when it is in the most naive way, to prove a theorem. It can be thought of as a primitive step of creating artificial thought processor. Any particular system can be thought as a geometric structure. Data which we sense by means of our sensory organs are the different `dimensions'. When we think we find out the relationship among different dimensions.

          \begin{figure}[t]
     \begin{center}
     \includegraphics[width=.35\textwidth]{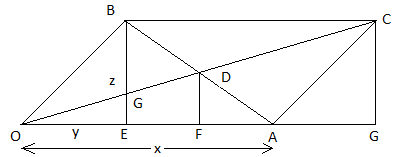}
     \end{center}
     \caption{A geometric system of a parallelogram.\label{f8}}
     \end{figure}

However, there are couples of challenges which we need to face while accomplishing an automated theorem prover in above mentioned method. They are listed below:
\begin{enumerate}
\item How the automated system would recognize which particular dimensions are required to be worked out to reach the goal. There are lots of dimensions possible, which we have ignored. For example, we have completely ignored the point of intersection between $OD$ and $BE$, say it is $G$ (\figurename~\ref{f8}). More dimensions like $OG, GD, BG$ and $GE$ are included. Unless we fix some heuristic constraints search space may get enormously enlarged. 
\item How the theorem prover would extract relationships among different dimensions extracting from the hypotheses. There should be a complete mechanism to do it.
\end{enumerate}

In this paper, we have discussed the overview of an automated theorem proving algorithm. While proving a theorem in Euclidian Logical inference method, the theorem prover should be skilled enough to inspect different portions of the geometric structure and to correlate them with the previously proven theorem(s), to infer useful decisions about different dimensions. It requires higher level of intelligence. At the very early stage, this is hard to accomplish. On the other hand, in Cartesian method lines and curves are represented by means of algebraic equations. It is done by following limited number of rules, hence more naive than the Euclidean method, resulting complicated calculations to solve the equations for some unknown variables. This method reduces the readability of the proof by increasing the complexity of calculations. Our proposed method assumes that the automated prover can 1) apply Pythagoras theorem and 2) apply the ratio of sides rule for similar triangles and can detect the situation where to apply them- this is an aspect of Euclidean logical inference method. A set of parameters will be defined by an expert and all other dimensions will be represented as functions of them similar to the Cartesian method. This primitive theorem prover shares aspects of both methods.
More research works are required to be performed to meet the requirements mentioned above to accomplish an automated geometric theorem prover resembling humane thought process.

\bibliographystyle{IEEEtran}
\bibliography{murtazapaper}

\end{document}